\documentclass{article} 
\usepackage{naaclhlt2016}
\usepackage{times}
\usepackage{latexsym}
\usepackage{hyperref}
\usepackage{url}
\usepackage{amsmath,amsthm,amsfonts}
\usepackage{multirow}
\usepackage{xspace}
\usepackage{tikz}
\usetikzlibrary{shapes,backgrounds,patterns}
\usepackage{graphicx}
\usepackage{subcaption}

\newcommand{\Prob}{\mathbb{P}}

\newcommand{\argOne}{\emph{arg1}\xspace}
\newcommand{\argTwo}{\emph{arg2}\xspace}

\newcommand{\citep}[1]{\cite{#1}}
\newcommand{\citet}[1]{\newcite{#1}}

\title{Multilingual Relation Extraction using Compositional Universal Schema}

\author{Patrick Verga, David Belanger, Emma Strubell, Benjamin Roth \& Andrew McCallum \\
College of Information and Computer Sciences\\
University of Massachusetts Amherst\\
\texttt{\{pat, belanger, strubell, beroth, mccallum\}@cs.umass.edu} \\
}

\begin{document}

\maketitle

\begin{abstract}
\emph{Universal schema} builds a knowledge base (KB) of entities and relations  by jointly embedding all relation types from input KBs as well as textual patterns expressing relations from raw text.
In most previous applications of universal schema, each textual pattern is represented as a single embedding, preventing generalization to unseen patterns. 
Recent work employs a neural network to capture patterns' compositional semantics, providing generalization to all possible input text.
In response, this paper introduces significant further improvements to the coverage and flexibility of universal schema relation extraction: predictions for entities unseen in training and multilingual transfer learning to domains with no annotation.
We evaluate our model through extensive experiments on the English and Spanish TAC KBP benchmark, outperforming the top system from TAC 2013 slot-filling using no handwritten patterns or additional annotation. 
We also consider a multilingual setting in which English training data entities overlap with the seed KB, but Spanish text does not. 
Despite having no annotation for Spanish data, we train an accurate predictor, with additional improvements obtained by tying word embeddings across languages. 
Furthermore, we find that multilingual training improves English relation extraction accuracy. 
Our approach is thus suited to broad-coverage automated knowledge base construction in a variety of languages and domains.
\end{abstract}


\section{Introduction}
\label{introduction}
The goal of automatic knowledge base construction (AKBC) is to build a structured knowledge base (KB) of facts using a noisy corpus of raw text evidence, and perhaps an initial seed KB to be augmented~\citep{NELL,yago,freebase}. AKBC supports downstream reasoning at a high level about extracted entities and their relations, and thus has broad-reaching applications to a variety of domains.

One challenge in AKBC is aligning knowledge from a structured KB with a text corpus in order to perform supervised learning through \emph{distant supervision}. \emph{Universal schema}~\citep{limin} along with its extensions~\citep{yao2013universal,vector_pra,neelakantan2015compositional,logicmfnaacl15}, avoids alignment by jointly embedding KB relations, entities, and surface text patterns. This propagates information between KB annotation and corresponding textual evidence.

The above applications of universal schema express each text relation as a distinct item to be embedded. This harms its ability to generalize to inputs not precisely seen at training time. Recently,~\citet{toutanova2015representing} addressed this issue by embedding text patterns using a deep sentence encoder, which captures the compositional semantics of textual relations and allows for prediction on inputs never seen before. 

\begin{figure}[h!]
\begin{center}
\vspace{-1.9069cm}

\def\firstcircle{(0,0) circle (1.5cm)}
\def\secondcircle{(0:2cm) circle (1.5cm)}
\def\midline{[line width=1pt, dashed] node[label={[label distance=-3cm]-15:in KB}] {} node[label={[label distance=-3.5cm]15:not in KB}] {} (-3,0) -- (5,0)}
\begin{tikzpicture}

	\begin{scope}
	  \clip \firstcircle (-3,3) rectangle (3,3);
	  \fill[pattern=north west lines, pattern color=black!70] \firstcircle;
	\end{scope}
	\begin{scope}
	  \clip \secondcircle (0,0) rectangle (4,4);
	  \fill[pattern=north east lines, pattern color=black!70] \secondcircle;
	\end{scope}
	\begin{scope}
	  \clip \firstcircle;
	  \clip (0.5,0) rectangle (3,2);
	  \fill[white] \secondcircle;
	\end{scope}
	\draw \firstcircle node[above=1.5cm] {English};
	\draw \secondcircle node[above=1.5cm] {Low-resource};
	\draw \midline;

\end{tikzpicture}
\caption{Splitting the entities in a multilingual AKBC training set into parts. We only require that entities in the two corpora overlap. Remarkably, we can train a model for the low-resource language even if entities in the low-resource language do not occur in the KB. \label{tab:multilingual-corpora}}
\end{center}
\vspace{-.4cm}
\end{figure}
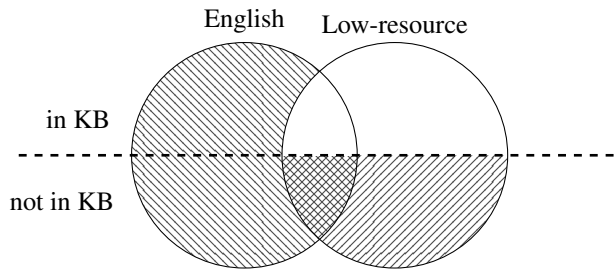

This paper further expands the coverage abilities of universal schema relation extraction by introducing techniques for forming predictions for new entities unseen in training and even for new domains with no associated annotation. In the extreme example of domain adaptation to a completely new language, we may have limited linguistic resources or labeled data such as treebanks, and only rarely a KB with adequate coverage. Our method performs multilingual transfer learning, providing a predictive model for a language with no coverage in an existing KB, by leveraging common representations for shared entities across text corpora. As depicted in Figure \ref{tab:multilingual-corpora}, we simply require that one language have an available KB of seed facts. We can further improve our models by tying a small set of word embeddings across languages using only simple knowledge about word-level translations, learning to embed semantically similar textual patterns from different languages into the same latent space. 

In extensive experiments on the TAC Knowledge Base Population (KBP) slot-filling benchmark we outperform the top 2013 system with an F1 score of 40.7 and perform relation extraction in Spanish with no labeled data or direct overlap between the Spanish training corpus and the training KB, demonstrating that our approach is well-suited for broad-coverage AKBC in low-resource languages and domains. Interestingly, joint training with Spanish improves English accuracy.

\section{Background \label{sec:background}}

AKBC extracts unary attributes of the form (\textit{subject}, \textit{attribute}), typed binary relations of the form (\textit{subject}, \textit{relation}, \textit{object}), or higher-order relations. We refer to subjects and objects as \textit{entities}. This work focuses solely on extracting binary relations, though many of our techniques generalize naturally to unary prediction. Generally, for example in Freebase~\citep{freebase}, higher-order relations are expressed in terms of collections of binary relations.

We now describe prior work on approaches to AKBC. They all aim to predict (\emph{s, r, o}) triples, but differ in terms of: (1) input data leveraged, (2) types of annotation required, (3) definition of relation label schema, and (4) whether they are capable of predicting relations for entities unseen in the training data. Note that all of these methods require pre-processing to detect entities, which may result in additional KB construction errors.

\subsection{Relation Extraction as Link Prediction \label{sec:prediction}}
A knowledge base is naturally described as a graph, in which entities are nodes and relations are labeled edges~\citep{yago,freebase}. In the case of \emph{knowledge graph completion}, the task is akin to link prediction, assuming an initial set of (\emph{s, r, o}) triples. See~\citet{nickel2015review} for a review. No accompanying text data is necessary, since links can be predicted using properties of the graph, such as transitivity. In order to generalize well, prediction is often posed as low-rank matrix or tensor factorization. A variety of model variants have been suggested, where the probability of a given edge existing depends on a multi-linear form~\citep{rescal,DBLP:journals/corr/Garcia-DuranBUG15,bishan,transe,wang2014knowledge,lin2015learning}, or non-linear interactions between $s$, $r$, and $o$~\citep{socherkb}.
Other approaches model the compositionality of multi-hop paths, typically for question answering \citep{bordes2014question,gu2015traversing,neelakantan2015compositional}.

\subsection{Relation Extraction as Sentence Classification
\label{seq:dist}}

Here, the training data consist of (1) a text corpus, and (2) a KB of seed facts with provenance, i.e. supporting evidence, in the corpus. Given individual an individual sentence, and pre-specified entities, a classifier predicts whether the sentence expresses a relation from a target schema. To train such a classifier, KB facts need to be aligned with supporting evidence in the text, but this is often challenging. For example, not all sentences containing Barack and Michelle Obama state that they are married. A variety of one-shot and iterative methods have addressed the alignment problem~\citep{bunescu2007learning,distant_supervision,riedel2010modeling,yao2010collective,hoffmann2011knowledge,surdeanu2012multi,min2013distant,zengdistant}.
An additional degree of freedom in these approaches is whether they classify individual sentences or predicting at the corpus level by aggregating information from all sentences containing a given pair of entities before prediction. The former approach is often preferable in practice, due to the simplicity of independently classifying individual sentences and the ease of associating each prediction with a provenance.
Prior work has applied deep learning to small-scale relation extraction problems, where functional relationships are detected between common nouns \citep{li2015tree,dos2015classifying}.
\citet{xu2015classifying} apply an LSTM to a parse path, while ~\citet{zengdistant} use a CNN on the raw text, with a special temporal pooling operation to separately embed the text around each entity.

\subsection{Open-Domain Relation Extraction
\label{sec:openIE}}
In the previous two approaches, prediction is carried out with respect to a fixed schema $R$ of possible relations $r$. This may overlook salient relations that are expressed in the text but do not occur in the schema. In response, \textit{open-domain} information extraction (OpenIE) lets the text speak for itself: $R$ contains all possible patterns of text occurring between entities $s$ and $o$~\citep{openie,etzioni2008open,resolver}. These are obtained by filtering and normalizing the raw text. The approach offers impressive coverage, avoids issues of distant supervision, and provides a useful exploratory tool. On the other hand, OpenIE predictions are difficult to use in downstream tasks that expect information from a fixed schema. 

Table~\ref{tab:patterns} provides examples of OpenIE patterns. The examples in row two and three illustrate relational contexts for which similarity is difficult to be captured by an OpenIE approach because of their syntactically complex constructions. This motivates the technique in Section~\ref{sec:encoder}, which uses a deep architecture applied to raw tokens, instead of rigid rules for normalizing text to obtain patterns.

\begin{table}[h!]
\small
\begin{center}
\begin{tabular}{|p{4.85cm}| p{2.37cm} | }
\hline
Sentence (context tokens italicized) & OpenIE pattern\\ \hline
{\bf Khan} \emph{'s younger sister,} {\bf Annapurna Devi}, who later married Shankar, developed into an equally accomplished master of the surbahar, but custom prevented her from performing in public. &  \argOne 's * sister \argTwo \\ \hline

A professor emeritus at Yale, {\bf Mandelbrot} \emph{was born in Poland but as a child moved with his family to} {\bf Paris} where he was educated. &  \argOne * moved with * family to \argTwo \\ \hline

{\bf Kissel} \emph{was born in Provo, Utah, but her family also lived in} {\bf Reno}. & \argOne * lived in \argTwo \\  \hline
\end{tabular}
\caption{Examples of sentences expressing relations. Context tokens (italicized) consist of the text occurring between entities (bold) in a sentence. OpenIE patterns are obtained by normalizing the context tokens using hand-coded rules. The top example expresses the per:siblings relation and the bottom two examples both express the per:cities\_of\_residence  relation. \label{tab:patterns}}
\end{center}
\vspace{-.4cm}
\end{table}

\subsection{Universal Schema}
When applying Universal Schema~\citep{limin} (USchema) to relation extraction, we combine the OpenIE and link-prediction perspectives.  By jointly modeling both OpenIE patterns and the elements of a target schema, the method captures broader relational structure than multi-class classification approaches that just model the target schema. Furthermore, the method avoids the distant supervision alignment difficulties of Section~\ref{seq:dist}. 

~\citet{limin} augment a knowledge graph from a seed KB with additional edges corresponding to OpenIE patterns observed in the corpus. Even if the user does not seek to predict these new edges, a joint model over all edges can exploit regularities of the OpenIE edges to improve modeling of the labels from the target schema. 

The data still consist of $(s,r,o)$ triples, which can be predicted using link-prediction techniques such as low-rank factorization.~\citet{limin} explore a variety of approximations to the 3-mode $(s,r,o)$ tensor. One such probabilistic model is:
\vspace{-.1cm}
\begin{equation}
\Prob \left((s,r,o)\right) = \sigma\left( u_{s,o}^\top v_r \right), \label{eq:US-prob}
\end{equation}
\vspace{-.5cm}

where  $\sigma()$ is a sigmoid function, $u_{s,o}$ is an embedding of the entity pair $(s,o)$, and $v_r$ is an embedding of the relation $r$, which may be an OpenIE pattern or a relation from the target schema. All of the exposition and results in this paper use this factorization, though many of the techniques we present later could be applied easily to the other factorizations described in ~\citet{limin}. Note that learning unique embeddings for OpenIE relations does not guarantee that similar patterns, such as the final two in Table~\ref{tab:patterns}, will be embedded similarly.

As with most of the techniques in Section~\ref{sec:prediction}, the data only consist of positive examples of edges. The absence of an annotated edge does not imply that the edge is false. In fact, we seek to predict some of these missing edges as true. ~\citet{limin} employ the Bayesian Personalized Ranking (BPR) approach of~\citet{rendle2009bpr}, which does not explicitly model unobserved edges as negative, but instead seeks to rank the probability of observed triples above unobserved triples.

Recently, \citet{toutanova2015representing} extended USchema to not learn individual pattern embeddings $v_r$, but instead to embed text patterns using a deep architecture applied to word tokens. This shares statistical strength between OpenIE patterns with similar words. We leverage this approach in Section~\ref{sec:encoder}. Additional work has modeled the regularities of multi-hop paths through knowledge graph augmented with text patterns~\citep{pra,pra_second,vector_pra,neelakantan2015compositional}.

\subsection{Multilingual Embeddings \label{sec:background-multilingual}}
Much work has been done on multilingual word embeddings.
Most of this work uses aligned sentences from the Europarl dataset~\citep{koehn2005europarl} to align word embeddings across languages~\citep{Gouws2015,luong2015bilingual,hermann2014multilingual}.
Others~\citep{mikolov2013,faruqui2014retrofitting} align separate single-language embedding models using a word-level dictionary.
\citet{mikolov2013} use translation pairs to learn a linear transform from one embedding space to another.

However, very little work exists on multilingual relation extraction. \citet{faruqui2015multilingual} perform multilingual OpenIE relation extraction by projecting all languages to English using Google translate. However, as explained in Section~\ref{sec:openIE} the OpenIE paradigm is not amenable to prediction within a fixed schema. Further, their approach does not generalize to low-resource languages where translation is unavailable -- while we use translation dictionaries to improve our results, our experiments demonstrate that our method is effective even without this resource.

%

\section {Methods}

\begin{figure*}[t!]
\caption{
Universal Schema jointly embeds KB and textual relations from Spanish and English, learning dense representations for entity pairs and relations using matrix factorization. Cells with a 1 indicate triples observed during training (left). The bold score represents a test-time prediction by the model (right). Using transitivity through KB/English overlap and English/Spanish overlap, our model can predict that a text pattern in Spanish evidences a KB relation despite no overlap between Spanish/KB entity pairs.
At train time we use BPR loss to maximize the inner product of entity pairs with KB relations and text patterns encoded using a bidirectional LSTM. At test time we score compatibility between embedded KB relations and encoded textual patterns using cosine similarity. In our Spanish model we treat embeddings for a small set of English/Spanish translation pairs as a single word, e.g. casado and married.
\label{fig:model}}
  \hspace{-.55cm}%
\begin{subfigure}{.7\textwidth}
    \includegraphics[scale=.65]{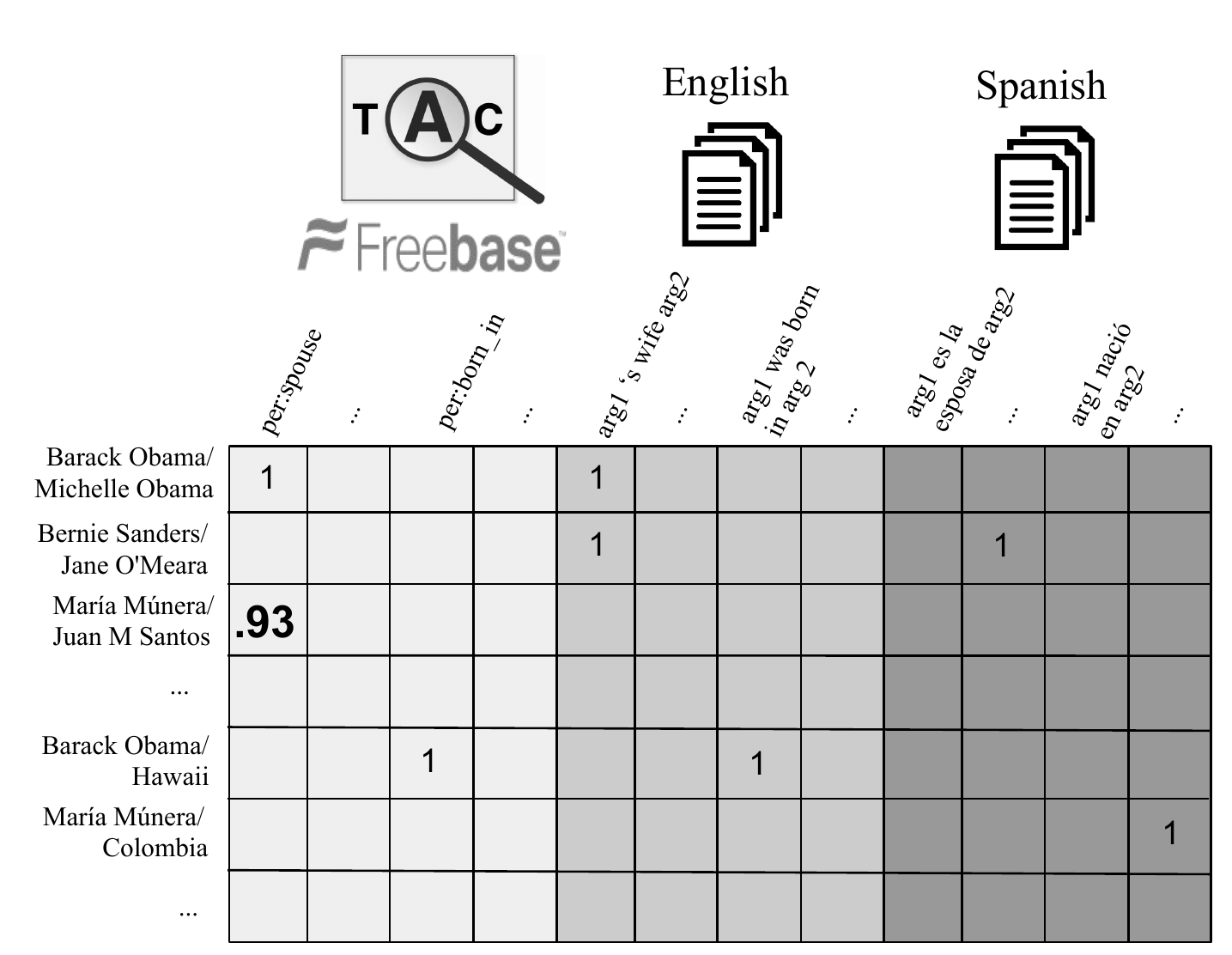}
\end{subfigure}%
  \hspace{-1.5cm}%
\begin{subfigure}{.4\textwidth}
  \includegraphics[scale=.75]{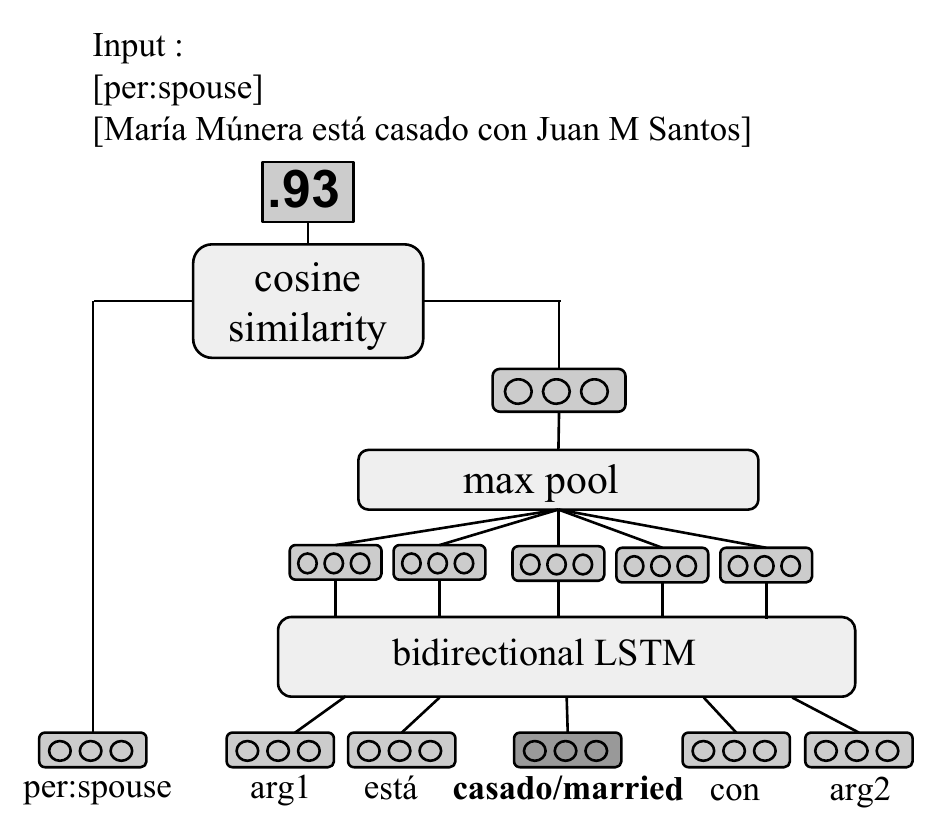}
\end{subfigure}
\end{figure*}

\subsection{Universal Schema as Sentence Classifier \label{sec:uschema}}
Similar to many link prediction approaches,~\citep{limin} perform transductive learning, where a model is learned jointly over train and test data. Predictions are made by using the model to identify edges that were unobserved in the test data but likely to be true. The approach is vulnerable to the \emph{cold start} problem in collaborative filtering~\citep{schein2002methods}: it is unclear how to form predictions for unseen entity pairs, without re-factorizing the entire matrix or applying heuristics.

In response, this paper re-purposes USchema as a means to train a sentence-level relation classifier, like those in Section~\ref{seq:dist}. This allows us to avoid errors from aligning distant supervision to the corpus, but is more deployable for real world applications. It also provides opportunities in Section~\ref{sec:multilingual} to improve multilingual AKBC.

We produce predictions using a very simple approach: (1) scan the corpus and extract a large quantity of triplets $(s,r_{\text{text}},o)$, where $r_{\text{text}}$ is an OpenIE pattern. For each triplet, if the similarity between the embedding of $r_{\text{text}}$ and the embedding of a target relation $r_{\text{schema}}$ is above some threshold, we predict the triplet $(s,r_{\text{schema}},o)$, and its provenance is the input sentence containing $(s,r_{\text{text}},o)$. We refer to this technique as~\textit{pattern scoring}. In our experiments, we use the cosine distance between the vectors (Figure \ref{fig:model}).
In Section~\ref{app:cosine}, we discuss details for how to make this distance well-defined.

\subsection{Using a Compositional Sentence Encoder to Predict Unseen Text Patterns \label{sec:encoder}}
The pattern scoring approach is subject to an additional cold start problem: input data may contain patterns unseen in training. This section describes a method for using USchema to train a relation classifier that can take arbitrary context tokens (Section~\ref{sec:openIE}) as input.

Fortunately, the cold start problem for context tokens is more benign than that of entities since we can exploit statistical regularities of text: similar sequences of context tokens should be embedded similarly. Therefore, following \citet{toutanova2015representing}, we  embed raw context tokens compositionally using a deep architecture. Unlike~\citet{limin}, this requires no manual rules to map text to OpenIE patterns and can embed any possible input string. The modified USchema likelihood is:
\begin{equation}
\Prob \left((s,r,o)\right) = \sigma\left( u_{s,o}^\top \text{Encoder}(r) \right).
\end{equation}
Here, if $r$ is raw text, then $\text{Encoder}(r)$ is parameterized by a deep architecture. If $r$ is from the target schema, $\text{Encoder}(r)$ is a produced by a lookup table (as in traditional USchema). Though such an encoder increases the computational cost of test-time prediction over straightforward pattern matching, evaluating a deep architecture can be done in large batches in parallel on a GPU.

Both convolutional networks (CNNs) and recurrent networks (RNNs) are reasonable encoder architectures, and we consider both in our experiments. CNNs have been useful in a variety of NLP applications~\citep{collobert2011natural,KalchbrennerACL2014,kim2014convolutional}. Unlike~\citet{toutanova2015representing}, we also consider RNNs, specifically Long-Short Term Memory Networks (LSTMs)~\citep{lstm}. LSTMs have proven successful in a variety of tasks requiring encoding sentences as vectors~\citep{rnnmt,rnnparse}. In our experiments, LSTMs outperform CNNs.

There are two key differences between our sentence encoder and that of~\citet{toutanova2015representing}.  First, we use the encoder at test time, since we process the context tokens for held-out data. On the other hand,~\citet{toutanova2015representing} adopt the transductive approach where the encoder is only used to help train better representations for the relations in the target schema; it is ignored when forming predictions.  Second, we apply the encoder to the raw text between entities, while~\citet{toutanova2015representing} first perform syntactic dependency parsing on the data and then apply an encoder to the path between the two entities in the parse tree. We avoid parsing, since we seek to perform multilingual AKBC, and many languages lack linguistic resources such as treebanks. Even parsing non-newswire English text, such as tweets, is extremely challenging. 

\subsection{Modeling Frequent Text Patterns}
\label{sec:non-comp}

Despite the coverage advantages of using a deep sentence encoder, separately embedding each OpenIE pattern, as in~\citet{limin}, has key advantages. In practice, we have found that many high-precision patterns occur quite frequently. For these, there is sufficient data to model them with independent embeddings per pattern, which imposes minimal inductive bias on the relationship between patterns. Furthermore, some discriminative phrases are idiomatic, i.e.. their meaning is not constructed compositionally from their constituents. For these, a sentence encoder may be inappropriate.

Therefore, pattern embeddings and deep token-based encoders have very different strengths and weaknesses. One values specificity, and models the head of the text distribution well, while the other has high coverage and captures the tail. In experimental results, we demonstrate that an ensemble of both models performs substantially better than either in isolation.


\subsection{Multilingual Relation Extraction with Zero Annotation \label{sec:multilingual}}

The models described in previous two sections provide broad-coverage relation extraction that can generalize to all possible input entities and text patterns, while avoiding error-prone alignment of distant supervision to a corpus. Next, we describe techniques for an even more challenging generalization task: relation classification for input sentences in completely different languages.

Training a sentence-level relation classifier, either using the alignment-based techniques of Section~\ref{seq:dist}, or the alignment-free method of Section~\ref{sec:uschema}, requires an available KB of seed facts that have supporting evidence in the corpus.  Unfortunately, available KBs have low overlap with corpora in many languages, since KBs have cultural and geographical biases. In response, we perform multilingual relation extraction by jointly modeling a high-resource language, such as English, and an alternative language with no KB annotation. This approach provides transfer learning of a predictive model to the alternative language, and generalizes naturally to modeling more languages.

Extending the training technique of Section~\ref{sec:uschema} to corpora in multiple languages can be achieved by factorizing a matrix that mixes data from a KB and from the two corpora. In Figure~\ref{tab:multilingual-corpora} we split the entities of a multilingual training corpus into sets depending on whether they have annotation in a KB and what corpora they appear in. We can perform transfer learning of a relation extractor to the low-resource language if there are entity pairs occurring in the two corpora, even if there is no KB annotation for these pairs. Note that we do not use the entity pair embeddings at test time: They are used only to bridge the languages during training. To form predictions in the low-resource language, we can simply apply the pattern scoring approach of Section~\ref{sec:uschema}.

In Section~\ref{sec:results}, we demonstrate that jointly learning models for English and Spanish, with no annotation for the Spanish data, provides fairly accurate Spanish AKBC, and even improves the performance of the English model. Note that we are not performing \textit{zero-shot} learning of a Spanish model~\citep{zeroshot}. The relations in the target schema are language-independent concepts, and we have supervision for these in English.

\subsection{Tied Sentence Encoders \label{sec:tie-words}}
The sentence encoder approach of Section~\ref{sec:encoder} is complementary to our multilingual modeling technique: we simply use a separate encoder for each language.  This approach is sub-optimal, however, because each sentence encoder will have a separate matrix of word embeddings for its vocabulary, despite the fact that there may be considerable shared structure between the languages. In response, we propose a straightforward method for tying the parameters of the sentence encoders across languages.

Drawing on the dictionary-based techniques described in Section \ref{sec:background-multilingual}, we first obtain a list of word-word translation pairs between the languages using a translation dictionary. The first layer of our deep text encoder consists of a word embedding lookup table. For the aligned word types, we use a single cross-lingual embedding.
Details of our approach are described in Appendix~\ref{sec:word-tying}.

\section{Task and System Description}

We focus on the TAC KBP slot-filling task. Much related work on embedding knowledge bases evaluates on the FB15k dataset \citep{transe,wang2014knowledge,lin2015learning,bishan,toutanova2015representing}. Here, relation extraction is posed as link prediction on a subset of Freebase.  This task does not capture the particular difficulties we address: (1) evaluation on entities and text unseen during training, and (2) zero-annotation learning of a predictor for a low-resource language.

Also, note both~\citet{toutanova2015representing} and~\citet{limin} explore the pros and cons of learning embeddings for entity pairs vs. separate embeddings for each entity. As this is orthogonal to our contributions, we only consider entity pair embeddings, which performed best in both works when given sufficient data.


\subsection{TAC Slot-Filling Benchmark}

The aim of the TAC benchmark is to improve both coverage and quality of relation extraction evaluation compared to just checking the extracted facts against a knowledge base, which can be incomplete and where the provenances are not verified. In the slot-filling task, each system is given a set of paired query entities and relations or `slots' to fill, and the goal is to correctly fill as many slots as possible along with provenance from the corpus. For example, given the query entity/relation pair (\emph{Barack Obama, per:spouse}), the system should return the entity \emph{Michelle Obama} along with sentence(s) whose text expresses that relation. The answers returned by all participating teams, along with a human search (with timeout), are judged manually for correctness, i.e. whether the provenance specified by the system indeed expresses the relation in question.


In addition to verifying our models on the 2013 and 2014 English slot-filling task, we evaluate our Spanish models on the 2012 TAC Spanish slot-filling evaluation. Because this TAC track was never officially run, the coverage of facts in the available annotation is very small, resulting in many correct predictions being marked incorrectly as precision errors. In response, we manually annotated all results returned by the models considered in Table~\ref{es-tac-table}. Precision and recall are calculated with respect to the union of the TAC annotation and our new labeling\footnote{Following \citet{surdeanu2012multi} we remove facts about undiscovered entities to correct for recall.}.


\subsection{Retrieval Pipeline \label{sec:pipeline}}
Our retrieval pipeline first generates all valid slot filler candidates for each query entity and slot, based on entities extracted from the corpus using {\sc Factorie} ~\citep{mccallum09:factorie:} to perform tokenization, segmentation, and entity extraction. We perform entity linking by heuristically linking all entity mentions from our text corpora to a Freebase entity using anchor text in Wikipedia. Making use of the fact that most Freebase entries contain a link to the corresponding Wikipedia page, we link all entity mentions from our text corpora to a Freebase entity by the following process:
First, a set of candidate entities is obtained by following frequent link anchor text statistics.
We then select that candidate entity for which the cosine similarity between the respective Wikipedia and the sentence context of the mention is highest, and link to that entity if a threshold is exceeded.

An entity pair qualifies as a candidate prediction if it meets the type criteria for the slot.\footnote{Due to the difficulty of retrieval and entity detection, the maximum recall for predictions is limited. For this reason, \citet{surdeanu2012multi} restrict the evaluation to answer candidates returned by their system and effectively rescaling recall. We do not perform such a re-scaling in our English results in order to compare to other reported results. Our Spanish numbers are rescaled. All scores reflect the `anydoc' (relaxed) scoring to mitigate penalizing effects for systems not included in the evaluation pool.} The TAC 2013 English and Spanish newswire corpora each contain about 1 million newswire documents from 2009--2012. The document retrieval and entity matching components of our relation extraction pipeline are based on RelationFactory~\citep{roth2014relationfactory}, the top-ranked system of the 2013 English slot-filling task. We also use the English distantly supervised training data from this system, which aligns the TAC 2012 corpus to Freebase.
 More details on alignment are described in Appendix \ref{sec:ds-el}.


As discussed in Section~\ref{sec:non-comp}, models using a deep sentence encoder and using a pattern lookup table have complementary strengths and weaknesses. In response, we present results where we ensemble the outputs of the two models by simply taking the union of their individual outputs. Slightly higher results might be obtained through more sophisticated ensembling schemes. 

\subsection {Model Details \label{sec:models}}
All models are implemented in Torch (code publicly available\footnote{\url{https://github.com/patverga/torch-relation-extraction}}).
Models are tuned to maximize F1 on the 2012 TAC KBP slot-filling evaluation.
We additionally tune the thresholds of our pattern scorer on a per-relation basis to maximize F1 using 2012 TAC slot-filling for English and the 2012 Spanish slot-filling development set for Spanish.
As in~\citet{limin}, we train using the BPR loss of~\citet{rendle2009bpr}.
Our CNN is implemented as described in \citet{toutanova2015representing}, using width-3 convolutions, followed by tanh and max pool layers.
The LSTM uses a bi-directional architecture where the forward and backward representations of each hidden state are averaged, followed by max pooling over time.
See Section \ref{sec:details}

We also report results including an alternate names (AN) heuristic, which uses automatically-extracted rules to detect the TAC `alternate name' relation.
To achieve this, we collect frequent Wikipedia link anchor texts for each query entity.
If a high probability anchor text co-occurs with the canonical name of the query in the same document, we return the anchor text as a slot filler.

\section{Experimental Results\label{sec:results}}

In experiments on the English and Spanish TAC KBC slot-filling tasks, we find that both USchema and LSTM models outperform the CNN across languages, and that the LSTM tends to perform slightly better than USchema as the only model. Ensembling the LSTM and USchema models further increases final F1 scores in all experiments, suggesting that the two different types of model compliment each other well. Indeed, in Section \ref{sec:uschema-lstm} we present quantitative and qualitative analysis of our results which further confirms this hypothesis:  the LSTM and USchema models each perform better on different pattern lengths and are characterized by different precision-recall tradeoffs.

\subsection {English TAC Slot-filling Results}

\begin{table}[t!]
\setlength{\tabcolsep}{4.1pt}
\begin{center}
\begin{tabular}{|lrrr|}
\hline
\bf Model & \bf Recall & \bf Precision & \bf F1 \\
\hline\hline
CNN                 & 31.6 & 36.8 & 34.1 \\
LSTM                & 32.2 & 39.6 & \bf 35.5  \\
USchema             & 29.4 & 42.6 & 34.8 \\
\hline\hline
USchema+LSTM        & 34.4 & 41.9 & 37.7 \\
USchema+LSTM+Es        & 38.1 & 40.2 & \bf 39.2 \\
\hline\hline
USchema+LSTM+AN	& 36.7 & 43.1 & 39.7 \\
USchema+LSTM+Es+AN & 40.2 & 41.2 & \bf 40.7 \\
\citet{roth2014relationfactory} & 35.8 & 45.7 & 40.2 \\

\hline
\end{tabular}
\caption{Precision, recall and F1 on the English TAC 2013 slot-filling task. AN refers to alternative names heuristic and Es refers to the addition of Spanish text at train time. LSTM+USchema ensemble outperforms any single model, including the highly-tuned top 2013 system of \protect\citet{roth2014relationfactory}, despite using no handwritten patterns.  
\label{en-tac-table}}
\end{center}
\vspace{-.3cm}
\end{table}
\begin{table}[t!]
\begin{center}
\begin{tabular}{|lrrr|}
\hline
\bf Model & \bf Recall & \bf Precision & \bf F1 \\
\hline\hline
CNN                 & 28.1 & 29.0 & 28.5 \\
LSTM                & 27.3 & 32.9 & \bf 29.8  \\
USchema             & 24.3 & 35.5 & 28.8 \\
\hline\hline
USchema+LSTM        & 34.1 & 29.3 & 31.5 \\
USchema+LSTM+Es        & 34.4 & 31.0 & \bf 32.6 \\

\hline
\end{tabular}
\caption{Precision, recall and F1 on the English TAC 2014 slot-filling task. Es refers to the addition of Spanish text at train time. The AN heuristic is ineffective on 2014 adding only 0.2 to F1. Our system would rank 4/18 in the official TAC 2014 competition behind systems that use hand-written patterns and active learning despite our system using neither of these additional annotations \protect\citep{SurdeanuMihai2014}.\label{2014-en-tac-table}}
\end{center}
\vspace{-.4cm}
\end{table}

Tables \ref{en-tac-table} and \ref{2014-en-tac-table} present the performance of our models on the 2013 and 2014 English TAC slot-filling tasks. Ensembling the LSTM and USchema models improves F1 by 2.2 points for 2013 and 1.7 points for 2014 over the strongest single model on both evaluations, LSTM. Adding the alternative names (AN) heuristic described in Section \ref{sec:models} increases F1 by an additional 2 points on 2013, resulting in an F1 score that is competitive with the state-of-the-art. We also demonstrate the effect of jointly learning English and Spanish models on English slot-filling performance. Adding Spanish data improves our F1 scores by 1.5 points on 2013 and 1.1 on 2014 over using English alone. This places are system higher than the top performer at the 2013 TAC slot-filling task even though our system uses no hand-written rules.

The state of the art systems on this task all rely on matching handwritten patterns to find additional answers while our models use only automatically generated, indirect supervision; even our AN heuristics (Section \ref{sec:pipeline}) are automatically generated. The top two 2014 systems were \citet{angeli2014stanford} and RPI Blender \citep{SurdeanuMihai2014} who achieved F1 scores of 39.5 and 36.4 respectively. Both of these systems used additional active learning annotation. The third place team \citep{Lin2014} relied on highly tuned patterns and rules and achieved an F1 score of 34.4. 

Our model performs substantially better on 2013 than 2014 for two reasons. First, our RelationFactory~\citep{roth2014relationfactory} retrieval pipeline was a top retrieval pipeline on the 2013 task, but was outperformed on the 2014 task which introduced new challenges such as confusable entities. Second, improved training using active learning gave the top 2014 systems a boost in performance. No 2013 systems, including ours, use active learning. \citet{bentortac14}, the 4th place team in the 2014 evaluation, used the same retrieval pipeline \citep{roth2014relationfactory} as our model and achieved an F1 score of 32.1.


\subsection{Spanish TAC Slot-filling Results \label{sec:qual-anal}}

\begin{table}
\begin{center}
\begin{tabular}{|lrrr|}
\hline
\bf Model & \bf Recall & \bf Precision & \bf F1 \\
\hline\hline
LSTM 		      &  9.3 & 12.5   & 10.7   \\
LSTM+Dict	      &  14.7 & 15.7  & 15.2   \\
USchema           &  15.2 & 17.5  & 16.3  \\
\hline\hline
USchema+LSTM       & 21.7 & 14.5  & 17.3  \\
USchema+LSTM+Dict  & 26.9 & 15.9  & \bf 20.0 \\
\hline
\end{tabular}
\caption{Zero-annotation transfer learning F1 scores on 2012 Spanish TAC KBP slot-filling task. Adding a translation dictionary improves all encoder-based models. Ensembling LSTM and USchema models performs the best. \label{es-tac-table}}
\end{center}
\vspace{-.6cm}
\end{table}


Table \ref{es-tac-table} presents 2012 Spanish TAC slot-filling results for our multilingual relation extractors trained using zero-annotation transfer learning. Tying word embeddings between the two languages results in substantial improvements for the LSTM. We see that ensembling the non-dictionary LSTM with USchema gives a slight boost over USchema alone, but ensembling the dictionary-tied LSTM with USchema provides a significant increase of nearly 4 F1 points over the highest-scoring single model, USchema. Clearly, grounding the Spanish data using a translation dictionary provides much better Spanish word representations. These improvements are complementary to the baseline USchema model, and yield impressive results when ensembled.

In addition to embedding semantically similar phrases from English and Spanish to have high similarity, our models also learn high-quality multilingual word embeddings. In Table \ref{joint-word} we compare Spanish nearest neighbors of English query words learned by the LSTM with dictionary ties versus the LSTM with no ties, using no unsupervised pre-training for the embeddings. Both approaches jointly embed Spanish and English word types, using shared entity embeddings, but the dictionary-tied model learns qualitatively better multilingual embeddings. 

\begin{table}[h]
\setlength{\tabcolsep}{3pt}
\small
\begin{center}
\begin{tabular}{|ll|}
\hline
\multicolumn{2}{|c|}{ \bf CEO}\\
\multicolumn{1}{|c}{Dictionary} & \multicolumn{1}{c|} {No Ties} \\ \hline 
jefe (chief)    & CEO \\ 
CEO & director (principle) \\
ejecutivo (executive)   &  directora (director) \\
cofundador (co-founder)  & firma (firm) \\
president (chairman) & magnate (tycoon)\\
\hline
\multicolumn{2}{|c|}{\bf headquartered}\\
\multicolumn{1}{|c}{Dictionary} & \multicolumn{1}{c|} {No Ties} \\ \hline
sede (headquarters) & Geol\'{o}gico (Geological) \\
situado (located) & Treki (Treki) \\
selectivo (selective) & Geof\'{i}sico(geophysical) \\
profesional (vocational) & Normand\'{i}a (Normandy)\\
bas\'{a}ndose (based) & emplea (uses)\\
\hline
\multicolumn{2}{|c|}{\bf hubby}\\
\multicolumn{1}{|c}{Dictionary} & \multicolumn{1}{c|} {No Ties} \\ \hline 
matrimonio (marriage)  & esposa (wife) \\ 
casada (married) & esposo (husband) \\
esposa (wife) &  casada(married) \\
cas\'{o} (married) & embarazada (pregnant)  \\
embarazada (pregnant) & embarazo (pregnancy) \\
\hline

\multicolumn{2}{|c|}{\bf alias}\\
\multicolumn{1}{|c}{Dictionary} & \multicolumn{1}{c|} {No Ties} \\ \hline
simplificado (simplified) & Weaver (Weaver)\\ 
sabido (known) & interrogaci\'{o}n (question) \\
seud\'{o}nimo (pseudonym)  &  alias \\
privatizaci\'{o}n (privatization)  & reelecto (reelected) \\
nombre (name)  & conocido (known)\\
\hline
\end{tabular}
\caption{Example English query words (not in translation dictionary) in bold with their top nearest neighbors by cosine similarity listed for the dictionary and no ties LSTM variants. Dictionary-tied nearest neighbors are consistently more relevant to the query word than untied. \label{joint-word}}
\end{center}
\end{table}

\subsection{USchema vs LSTM \label{sec:uschema-lstm}}

\begin{figure}[t!]
\begin{center}
\includegraphics[scale=0.45]{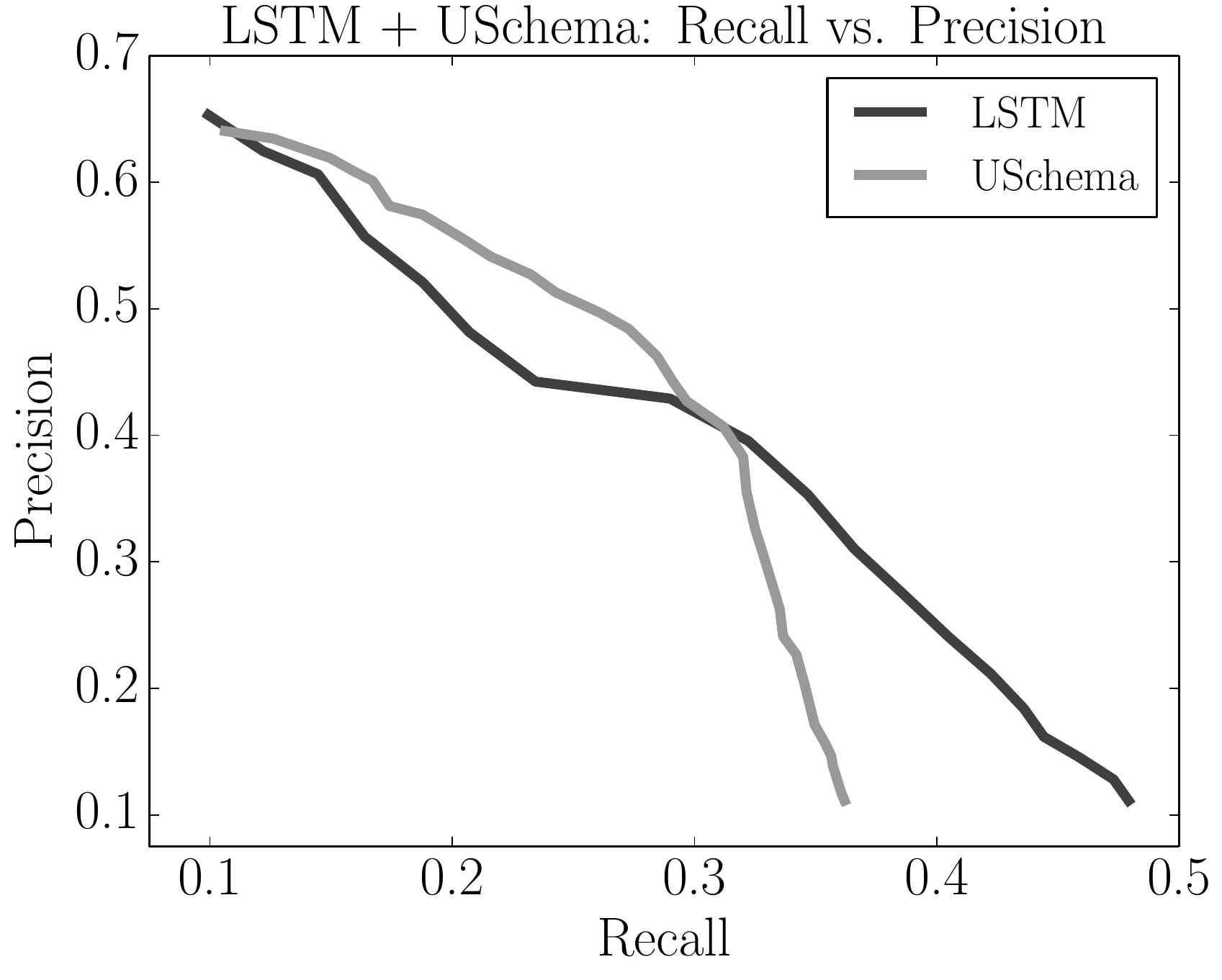}
\caption{Precision-Recall curves for USchema and LSTM on 2013 TAC slot-filling. USchema achieves higher precision values whereas LSTM has higher recall. \label{fig:pr-curve}}
\end{center}
\end{figure}

\begin{figure}[t!]
\begin{center}
\includegraphics[scale=0.45]{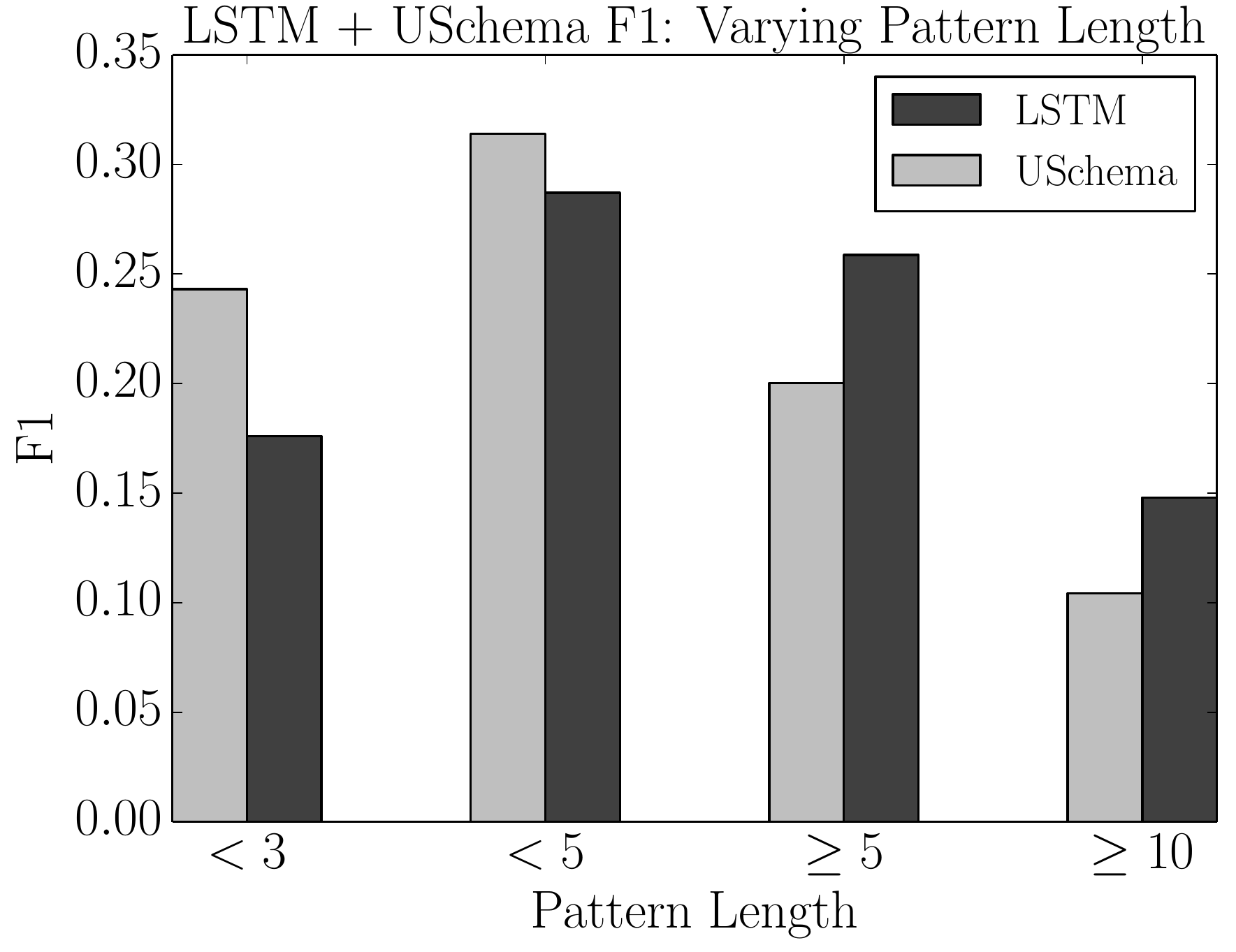}
\caption{F1 achieved by USchema vs. LSTM models for varying pattern token lengths on 2013 TAC slot-filling. LSTM performs better on longer patterns whereas USchema performs better on shorter patterns. \label{fig:f1-vary-pats}}
\end{center}
\vspace{-.4cm}
\end{figure}

We further analyze differences between USchema and LSTM in order to better understand why ensembling the models results in the best performing system. Figure \ref{fig:pr-curve} depicts precision-recall curves for the two models on the 2013 slot-filling task. As observed in earlier results, the LSTM achieves higher recall at the loss of some precision, whereas USchema can make more precise predictions at a lower threshold for recall. In Figure \ref{fig:f1-vary-pats} we observe evidence for these different precision-recall trade-offs: USchema scores higher in terms of F1 on shorter patterns whereas the LSTM scores higher on longer patterns. As one would expect, USchema successfully matches more short patterns than the LSTM, making more precise predictions at the cost of being unable to predict on patterns unseen during training. The LSTM can predict using any text between entities observed at test time, gaining recall at the loss of precision. Combining the two models makes the most of their strengths and weaknesses, leading to the highest overall F1.

Qualitative analysis of our English models also suggests that our encoder-based models (LSTM) extract relations based on a wide range of semantically similar patterns that the pattern-matching model (USchema) is unable to score due to a lack of exact string match in the test data. For example, Table \ref{tab:lstm-us-similar-rels} lists three examples of the \emph{per:children} relation that the LSTM finds which USchema does not, as well as three patterns that USchema does find. Though the LSTM patterns are all semantically and syntactically similar, they each contain different specific noun phrases, e.g. \emph{Lori}, \emph{four children}, \emph{toddler daughter}, \emph{Lee and Albert}, etc. Because these specific nouns weren't seen during training, USchema fails to find these patterns whereas the LSTM learns to ignore the specific nouns in favor of the overall pattern, that of a parent-child relationship in an obituary. USchema is limited to finding the relations represented by patterns observed during training, which limits the patterns matched at test-time to short and common patterns; all the USchema patterns matched at test time were similar to those listed in Table \ref{tab:lstm-us-similar-rels}: variants of \emph{'s son, '}.

\begin{table}[h]
\begin{center}
\small
\begin{tabular}{|p{7.6cm}|}
\hline
\multicolumn{1}{|c|}{\textbf{LSTM}} \\ \hline
{\bf McGregor} \emph{is survived by his wife, Lori, and four children, daughters Jordan,} { \bf Taylor} and Landri, and a son, Logan. \\ \hline
In addition to his wife, {\bf Mays} \emph{is survived by a toddler daughter and a son,} {\bf Billy Mays Jr.}, who is in his 20s. \\ \hline
{\bf Anderson} \emph{is survived by his wife Carol, sons Lee and Albert, daughter} {\bf Shirley Englebrecht} and nine grandchildren. \\
\hline\hline
\multicolumn{1}{|c|}{\textbf{USchema}}  \\ \hline
{\bf Dio} \emph{'s son,} {\bf Dan Padavona}, cautioned the memorial crowd to be screened regularly by a doctor and take care of themselves, something he said his father did not do. \\ \hline
But {\bf Marshall} \emph{'s son,} {\bf Philip}, told a different story.  \\ \hline
``I'd rather have Sully doing this than some stranger, or some hotshot trying to
be the next Billy Mays,'' said the guy who actually is the next {\bf Billy Mays}\emph{, his son} {\bf Billy Mays III}. \\
\hline
\end{tabular}
\caption{Examples of the \emph{per:children} relation discovered by the LSTM and Universal Schema. Entities are bold and patterns italicized. The LSTM models a richer set of patterns \label{tab:lstm-us-similar-rels}}
\end{center}
\vspace{-.5cm}
\end{table}

\section{Conclusion}

By jointly embedding English and Spanish corpora along with a KB, we can train an accurate Spanish relation extraction model using no direct annotation for relations in the Spanish data. This approach has the added benefit of providing significant accuracy improvements for the English model, outperforming the top system on the 2013 TAC KBC slot filling task, without using the hand-coded rules or additional annotations of alternative systems. By using deep sentence encoders, we can perform prediction for arbitrary input text and for entities unseen in training. Sentence encoders also provides opportunities to improve cross-lingual transfer learning by sharing word embeddings across languages. In future work we will apply this model to many more languages and domains besides newswire text. We would also like to avoid the entity detection problem by using a deep architecture to both identify entity mentions and identify relations between them.

\subsubsection*{Acknowledgments}
Many thanks to Arvind Neelakantan for good ideas and discussions. We also appreciate a generous hardware grant from nVidia. This work was supported in part by the Center for Intelligent Information Retrieval, in part by Defense Advanced Research Projects Agency (DARPA) under agreement \#FA8750-13-2-0020 and contract \#HR0011-15-2-0036, and in part by the National Science Foundation (NSF) grant numbers DMR-1534431, IIS-1514053 and CNS-0958392. The U.S. Government is authorized to reproduce and distribute reprints for Governmental purposes notwithstanding any copyright notation thereon, in part by DARPA via agreement \#DFA8750-13-2-0020 and NSF grant \#CNS-0958392. Any opinions, findings and conclusions or recommendations expressed in this material are those of the authors and do not necessarily reflect those of the sponsor.

\bibliography{sources}
\bibliographystyle{naaclhlt2016}

\newpage
\section{Appendix}

\subsection{Additional Qualitative Results \label{sec:more-qual-anal}}

Qualitative analysis of our multilingual models further suggests that they successfully embed semantically similar relations across languages using tied entity pairs and translation dictionary as grounding. Table \ref{tab:cross-lingual-relations} lists three top nearest neighbors in English for several Spanish patterns from the text. In each case, the English patterns capture the relation represented in the Spanish text.

\newcommand{\tablespace}{\end{tabular}
\newline
\newline
\begin{tabular}{|p{7.6cm}|}
}
\begin{table}[h]
\begin{center}
\small
\begin{tabular}{|p{7.6cm}|}
\hline
 \bf y cuatro de sus familias, incluidos su esposa, \endgraf \hspace{5pt}Wu Shu-chen, su hijo, \\
\it{ and four of his family members, including his wife, \endgraf \hspace{5pt}Wu Shu-chen, his son, } \\
\hline
 and his son  \\
 is survived by his wife, Sybil MacKenzie and a son,  \\
 gave birth to a baby last week -- son  \\
\hline
\tablespace
\hline
 \bf (Puff Daddy, cuyos verdaderos nombre sea  \\
\it{ (Puff Daddy, whose real name is } \\
\hline
 (usually credited as {\it E1} \\
 (also known as Gero \#\#, real name  \\
 and (after changing his name to  \\
\hline
\tablespace
\hline
 \bf lleg\'{o} a la alfombra roja en compa\~{n}\'{i}a de su esposa, la \endgraf \hspace{5pt} actriz  Suzy Amis, casi una hora antes que su ex esposa, \\
\it{ arrived on the red carpet with his wife, \endgraf \hspace{5pt} actress  Suzy Amis, nearly an hour before his ex-wife , } \\
\hline
, who may or may not be having twins with husband  \\
, aged twenty, Kirk married \\
 went to elaborate lengths to keep his wedding to former \endgraf \hspace{5pt}supermodel \\
\hline
\end{tabular}
\caption{Top English patterns for a Spanish query pattern encoded using the dictionary LSTM: For each Spanish query (English translation in italics), a list of English nearest neighbors. \label{tab:cross-lingual-relations}}
\end{center}
\end{table}

Our model jointly embeds KB relations together with English and Spanish text. We demonstrate that plausible textual patterns are embedded close to the KB relations they express. Table \ref{tab:top-tac-patterns} shows top scoring English and Spanish patterns given sample relations from our TAC KB.

\begin{table}[h]
\begin{center}
\begin{tabular}{|p{7.8cm}|}
\hline
\textbf{per:sibling} \\
\hline
   \argOne, seg\'{u}n petici\'{o}n the primeros ministro, \endgraf \hspace{5pt} su hermano gemelo \argTwo  			\\ 
  \argOne, sea the principal favorito para esto oficina \endgraf \hspace{5pt}que tambi\'{e}n ambiciona su hermano \argTwo 	\\
  \argOne, y su hermano gemelo, the primeros ministro \argTwo 	\\
\hline
  \argOne, for whose brother \argTwo  		\\
  \argOne inherited his brother \argTwo 	\\
  \argOne on saxophone and brother \argTwo 	\\
\hline\hline
\textbf{org:top\_members\_employees} \\
\hline
   \argTwo, presidente y director generales the \argOne  			\\
   	\argTwo, presidente of the negocios especializada \argOne  	\\
   	\argTwo (CIA), the director of the entidad, \argOne 	\\
\hline
 \argTwo, vice president and policy director of the \argOne  		\\
 \argTwo, president of the German Soccer \argOne 	\\
  \argTwo, president of the quasi-official \argOne 	\\
\hline\hline
\textbf{per:alternate\_names} \\
\hline
   \argOne(como tambi\'{e}n son sabido para \argTwo 			\\
   \argTwo-cuyos verdaderos nombre sea \argOne 	\\
   	\argOne  tambi\'{e}n sabido como \argTwo 	\\
\hline
   \argOne aka \argTwo 		\\
   \argOne, who also creates music under the pseudonym \argTwo 	\\
   \argOne( of Modern Talking fame ) aka \argTwo  	\\
\hline\hline
\textbf{per:cities\_of\_residence} \\
 \hline
  \argOne, poblado d\'{o}nde vive \argTwo 			\\
   \argOne, una ciudadano naturalizado american\endgraf \hspace{5pt} y nacido in \argTwo 	\\
   \argOne, que vive in \argTwo 	\\
\hline
   \argOne was born Jan. \# , \#\#\#\# in \argTwo 		\\
   	\argOne was born on Monday in \argTwo 	\\
   \argOne was born at Keighley in \argTwo 	\\
\hline
\end{tabular}
\caption{Top scoring patterns for both Spanish (top) and English (bottom) given query TAC relations. \label{tab:top-tac-patterns}}
\end{center}
\end{table}

\subsection {Implementation and Hyperparameters \label{sec:details}}
We performed a small grid search over learning rate {0.0001, 0.005, 0.001}, dropout {0.0, 0.1, 0.25, 0.5}, dimension {50, 100}, $\ell_2$ gradient clipping {1, 10, 50}, and epsilon {1e-8, 1e-6, 1e-4}. All models are trained for a maximum of 15 epochs. The CNN and LSTM both use 100d embeddings while USchema uses 50d. The CNN and LSTM both learned 100-dimensional word embeddings which were randomly initialized. Using pre-trained embeddings did not substantially affect the results. Entity pair embeddings for the baseline USchema model are randomly initialized. For the models with LSTM and CNN text encoders, entity pair embeddings are initialized using vectors from the baseline USchema model. This performs better than random initialization. We perform $\ell_2$ gradient clipping to 1 on all models. Universal Schema uses a batch size of 1024 while the CNN and LSTM use 128. All models are optimized using ADAM \citep{kingma2014adam} with $\epsilon=1e-8$, $\beta_1=0.9$, and $\beta_2=0.999$ with a learning rate of .001 for USchema and .0001 for CNN and LSTM. The CNN and LSTM also use dropout of 0.1 after the embedding layer.

\subsection{Details Concerning Cosine Similarity Computation \label{app:cosine}}
We measure the similarity between $r_{\text{text}}$ and $r_{\text{schema}}$ by computing the vectors' cosine similarity. However, such a distance is not well-defined, since the model was trained using inner products between entity vectors and relation vectors, not between two relation vectors. The US likelihood is invariant to invertible transformations of the latent coordinate system, since $\sigma\left( u_{s,o}^\top v_r \right) = \sigma\left( (A^\top u_{s,o})^\top A^{-1} v_r \right)$ for any invertible $A$. When taking inner products between two $v$ terms, however, the implicit $A^{-1}$ terms do not cancel out. We found that this issue can be minimized, and high quality predictive accuracy can be achieved, simply by using sufficient $\ell_2$ regularization to avoid implicitly learning an $A$ that substantially stretches the space.

\subsection{Data Pre-processing, Distant Supervision and Extraction Pipeline \label{sec:ds-el}}

We replace tokens occurring less than 5 times in the corpus with UNK and normalize all digits to \# (e.g. Oct-11-1988 becomes Oct-\#\#-\#\#\#\#).
For each sentence, we then extract all entity pairs and the text between them as surface patterns, ignoring patterns longer than 20 tokens.
This results in 48 million English `relations'. In Section~\ref{sec:norm}, we describe a technique for normalizing the surface patterns.
We filter out entity pairs that occurred less than 10 times in the data and extract the largest connected component in this entity co-occurrence graph.
This is necessary for the baseline US model, as otherwise learning decouples into independent problems per connected component.
Though the components are connected when using sentence encoders, we use only a single component to facilitate a fair comparison between modeling approaches.
We add the distant supervision training facts from the RelationFactory system, i.e. 352,236 entity-pair-relation tuples obtained from Freebase and high precision seed patterns.
The final training data contains a set of 3,980,164 (KB and openIE) facts made up of 549,760 unique entity pairs, 1,285,258 unique relations and 62,841 unique tokens.

We perform the same preprocessing on the Spanish data, resulting in 34 million raw surface patterns between entities.
We then filter patterns that never occur with an entity pair found in the English data.  This yields 860,502 Spanish patterns.
Our multilingual model is trained on a combination of these Spanish patterns, the English surface patterns, and the distant supervision data described above.
We learn word embeddings for 39,912 unique Spanish word types.
After parameter tying for translation pairs (Section \ref{sec:tie-words}),  there are 33,711 additional Spanish words not tied to English.

\subsection{Generation of Cross-Lingual Tied Word Types}
\label{sec:word-tying}
We follow the same procedure for generating translation pairs as \cite{mikolov2013}. First, we select the top 6000 words occurring in the lowercased Europarl dataset for each language and obtain a Google translation. We then filter duplicates and translations resulting in multi-word phrases. We also remove English past participles (ending in -ed) as we found the Google translation interprets these as adjectives (e.g.,  `she read the borrowed book' rather than `she borrowed the book') and much of the relational structure in language we seek to model is captured by verbs. This resulted in 6201 translation pairs that occurred in our text corpus. Though higher quality translation dictionaries would likely improve this technique, our experimental results show that such automatically generated dictionaries perform well.

\subsection{Open IE Pattern Normalization}
\label{sec:norm}
To improve US generalization, our US relations use log-shortened patterns where the middle tokens in patterns longer than five tokens are simplified. For each long pattern we take the first two tokens and last two tokens, and replace all $k$ remaining tokens with the number $\log k$. For example, the pattern {\bf Barack Obama} {\it is married to a person named} {\bf Michelle Obama} would be converted to: {\bf Barack Obama} {\it is married [1] person named} {\bf Michell Obama}. This shortening performs slightly better than whole patterns. LSTM and CNN variants use the entire sequence of tokens.

\end{document}